\documentclass{article}
\pdfoutput=1
\usepackage[preprint]{spconf1}
\usepackage{amsmath,graphicx}
\usepackage{amsfonts}
\usepackage[ruled]{algorithm2e}
\usepackage{algorithmic,algorithm2e}
\usepackage{booktabs}
\usepackage{multirow}
\usepackage{color}
\usepackage{floatrow}
\usepackage{marvosym}
\copyrightnotice{\begin{minipage}{\textwidth}
  \centering
  \copyright\ IEEE 2021. Personal use of this material is permitted. Permission from IEEE must be obtained for all other uses, in any current or future media, including reprinting/republishing this material for advertising or promotional purposes, creating new collective works, for resale or redistribution to servers or lists, or reuse of any copyrighted component of this work in other works.
  \end{minipage}
  }

\newfloatcommand{figurebox}{figure}[\nocapbeside][\dimexpr(\textwidth-\columnsep)/2\relax]
\newfloatcommand{tablebox}{table}[\nocapbeside][\dimexpr(\textwidth-\columnsep)/2\relax]

\title{Semi-Supervised 3D Hand-Object Pose Estimation \\
via Pose Dictionary Learning}
\name{Zida~Cheng, Siheng~Chen\textsuperscript{\Letter}, Ya~Zhang\textsuperscript{\Letter} \thanks{This work is supported by the National Key Research and Development Program of China (No. 2020YFB1406801), 111 plan (No. BP0719010),  and STCSM (No. 18DZ2270700), and State Key Laboratory of UHD Video and Audio Production and Presentation.}}

\address{Cooperative Medianet Innovation Center, Shanghai Jiao Tong University, Shanghai, China}

\begin{document}
\ninept
\maketitle

\begin{abstract}
3D hand-object pose estimation is an important issue to understand the interaction between human and environment. Current hand-object pose estimation methods require detailed 3D labels, which are expensive and labor-intensive. To tackle the problem of data collection, we propose a semi-supervised 3D hand-object pose estimation method with two key techniques: pose dictionary learning and an object-oriented coordinate system. The proposed pose dictionary learning module can distinguish infeasible poses by reconstruction error, enabling unlabeled data to provide supervision signals. The proposed object-oriented coordinate system can make 3D estimations equivariant to the camera perspective. Experiments are conducted on FPHA and HO-3D datasets. Our method reduces estimation error by 19.5\% / 24.9\% for hands/objects compared to straightforward use of labeled data on FPHA and outperforms several baseline methods. Extensive experiments also validate the robustness of the proposed method.

\end{abstract}
\section{Introduction}
\label{sec:intro}
 
Estimating the pose of hands and manipulated objects is an important task to understand and recognize the behavior of human beings. The related techniques are highly valuable for many practical applications, e.g., AR/VR games and robotics~\cite{FPHAafford,handpose_review,handpose_app,handseg_vr}. Here we specifically consider 3D hand-object pose estimation; that is, given the 2D poses of a pair of hand and object, we aim to estimate the corresponding 3D poses. Recent 3D hand-object pose estimation methods use deep neural networks (CNN~\cite{Tekin_2019_CVPR,Obman,HO_seperate}, GCN~\cite{HOPE}, transformer~\cite{HOT,transformer}) to jointly model hands and objects. However, these methods require detailed 3D labels, demanding expensive sensing equipment and a huge amount of manpower~\cite{FPHA,HO3D,contactpose}. To tackle the problem of data collection, we propose to perform 3D hand-object pose estimation in a~\emph{semi-supervised} manner; that is, we only label a small subset of hand-object poses and use both labeled and unlabeled data to train a pose estimator.

\begin{figure}[t!]
\centering
\includegraphics[width=7.6cm]{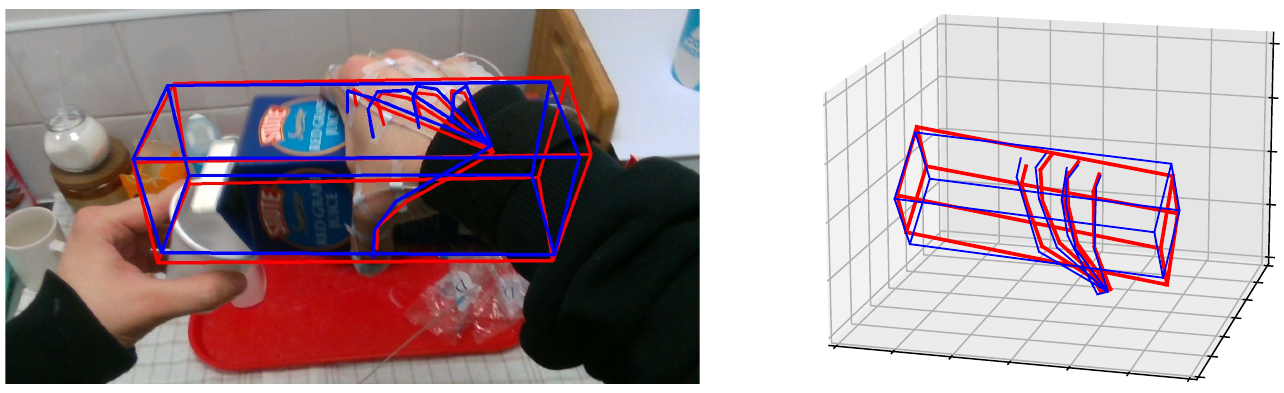}
\vspace{-15pt}
\caption{Hand-Object pose estimation results obtained by the proposed method from the camera view (left) and another view (right). Blue lines are ground truth. 
\vspace{-15pt}
}
\label{fig:res}
\end{figure}


There are two main challenges in semi-supervised 3D hand-object pose estimation task: (1) unlabeled data should be effectively leveraged to improve the estimation; and (2) the estimation result should be equivariant with respect to the camera perspective; that is, for multiple 2D poses obtained through photographing the same 3D pose from different camera perspectives, 3D estimation results should be consistent in terms of geometric shapes and only vary in terms of camera perspectives.

To address the first challenge, we introduce an internal self-supervision task and use labeled data to train a 3D grasping pose dictionary, which can well reconstruct feasible 3D grasping poses; in other words, this trained 3D grasping pose dictionary can use the reconstruction errors to distinguish feasible 3D grasping poses from infeasible ones, acting as a discriminator. In our system, even though we do not know the ground-truth 3D grasping poses for unlabeled data, we can use this 3D grasping pose dictionary to identify bad 3D estimations and supervise our pose estimation module to adjust. To address the second challenge, we propose an object-oriented cylindrical coordinate system, which uses the center of the object as the origin and synchronizes the object's orientation. In this coordinate system, the joints' positions are invariant to the camera perspective and an estimation model can focus on estimating grasping poses, excluding the influence of the camera's perspective. After the estimation, we can transform the result back to the standard camera-based Cartesian coordinate system, which makes the estimation equivariant to camera's perspective.

Integrating above designs, we propose a novel semi-supervised 3D hand-object pose estimation network, which includes two training phases. In the first phase, we use labeled data only to train a~\emph{pose dictionary learning} module, whose functionality is to decompose an input 3D grasping pose into a linear combination of 3D grasping pose atoms. The training process is to learn such a 3D grasping pose dictionary through self-reconstruction. Different from vanilla autoencoders~\cite{AE_anomaly_detection,AE_image_compression, AE_feature, AE_denoising}, we use the 3D grasping pose dictionary and the corresponding linear approximation to regularize the reconstruction process, whose benefits are to model the grasping poses more explicitly and constrain the variance of reconstruction process. In the second phase, we train our pose estimation module with the pose dictionary learning module fixed. The pose estimation module is implemented by a graph U-net~\cite{HOPE} model and is trained with both labeled and unlabeled data, where labeled data provides direct supervision and unlabeled data provides supervision signals through the reconstruction error of the pose dictionary learning module. In both phases, the proposed object-oriented cylindrical coordinate is used in the reconstruction process. We conduct experiments on FPHA~\cite{FPHA}  and HO-3D~\cite{HO3D} datasets and the experimental results show that our method improves estimation accuracy significantly and is of high robustness.

The main contributions of our method are as follows.

$\bullet$ We propose a semi-supervised 3D hand-object pose estimation network, which is trained by leveraging both labeled and unlabeled data. To our best knowledge, this is the first work to tackle 3D hand-object pose estimation task in a semi-supervised setting.

$\bullet$ We propose a pose dictionary learning module to perform an auxiliary self-reconstruction task, enabling unlabeled data to provide supervision signals.

$\bullet$  We propose an object-oriented cylindrical coordinate system to represent 3D poses, making 3D estimation results equivariant to the camera perspective.


$\bullet$ We conduct experiments on FPHA~\cite{FPHA} and HO-3D~\cite{HO3D} datasets. The proposed method reduces the estimation error by 19.5\%/24.9\% for hands/objects compared to straightforward use of labeled data on FPHA, and outperforms several baseline methods. Extensive experiments show that our method is robust to the pose dictionary size and weight of the reconstruction loss.

\section{Methodology}
\label{sec:format}

\subsection{Problem Formulation}
The goal of 3D hand-object pose estimation is to estimate the 3D coordinates of hand joints and the object bounding box's 8 corners from the corresponding 2D coordinates. Mathematically, let $ \mathcal{X} = \left\{\mathbf{X}_{i}\right\}_{i=1}^N$ be the 2D pose input, where $\mathbf{X}_{i} \in \mathbb{R}^{2 \times (m+8)}$ with $m$ is the number of joints in a hand. Among them, only $N_L$ samples $ \mathcal{X}_L = \left\{\mathbf{X}_{i}\right\}_{i=1}^{N_L}$, have 3D annotations; denoted as $ \mathcal{Y}_L = \left\{\mathbf{Y}_{i}\right\}_{i=1}^{N_L}$, where $\mathbf{Y}_{i} \in \mathbb{R}^{3 \times (m+8)}$. Let $\mathcal{P}_L = \left\{(\mathbf{X}_i, \mathbf{Y}_i)\right\}_{i=1}^{N_L} $ be the annotated pairs of 2D-3D poses. The rest $N - N_L$ samples $\mathcal{X}_U = \left\{\mathbf{X}_i\right\}_{i=N_L+1}^N$ do not have 3D annotations. The task is to infer 3D poses from 2D poses through the limited annotated pairs $\mathcal{P}_L$ as well as a huge amount of unannotated 2D poses $\mathcal{X}_U$.

Here we tackle this task by proposing a neural-network-based estimation model; that is, given both labeled and unlabeled data for training, we aim to propose a pose estimation network to generate 3D estimations for unlabeled data. Different from many other recent works~\cite{Tekin_2019_CVPR,Obman,HO_seperate,HOPE,HOT,HO_unknownobj,HO_feedback}, here we consider a semi-supervised setting; that is, besides limited labeled data, we are allowed to involve a huge amount of unlabeled data to train the network. To handle this new setting, our system includes two learning modules: a pose estimation module, which estimates 3D poses from 2D poses, and a dictionary learning module, which enables the supervision from unlabeled data. The whole training procedure has two phases. In the first phase, we train the pose dictionary learning module based on labeled data. In the second phase, we fix the pose dictionary learning module and train the pose estimation module based on both labeled and unlabeled data. Note that the pose dictionary learning module is an auxiliary module that benefits the training of unlabeled data. During the inference time, we only need the pose estimation module to provide 3D estimations.

\subsection{Object-Oriented Cylindrical Coordinate System}
\begin{figure}[t]
    \centering
    \includegraphics[width=6.0cm]{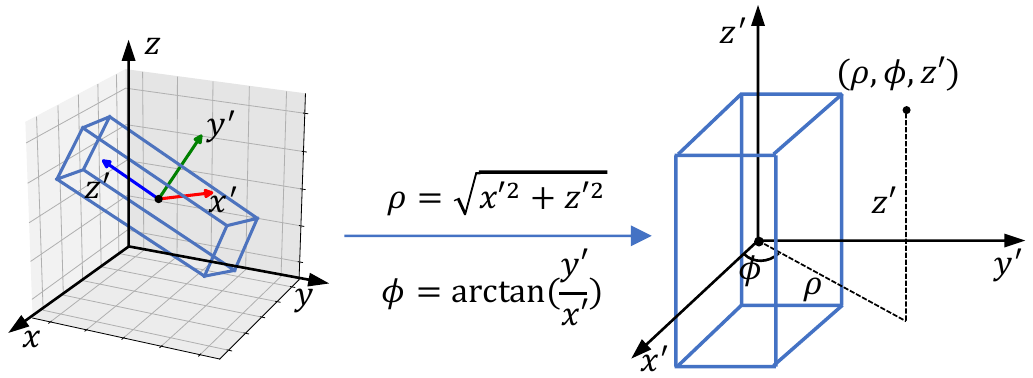}
    \vspace{-14pt}
    \caption{Transformation from camera coordinate to the proposed object-oriented cylindrical coordinate.
    \vspace{-14pt}
    }
    \label{fig:cordtrans}
\end{figure}

Before presenting our estimation model, we first introduce  a synchronized coordinate system to represent each 3D grasping pose. The new coordinate system can make the estimation result equivariant to the camera perspective. We firstly consider transforming the camera coordinate $xyz$ to an object-oriented Cartesian coordinate system $x'y'z'$ , where the origin is the center of the object bounding box and the axes are parallel to the edges. Then the proposed object-oriented cylindrical coordinate $(\rho,\phi,z')$ is defined based on $x'y'z'$. In this coordinate system, $\rho$ is the distance to the $z'$ axis and $\phi$ is the polar angle with respect to $x'$ axis. The whole transformation is illustrated in Fig. \ref{fig:cordtrans}. 

As the definition intrinsically lays objects at the center, we only consider the $m$ hand joints in the final cylindrical coordinates. Another problem is the numerical representation of $\phi$. For example, $- 0.99\pi $ and $0.99\pi$ are almost the same angle but their numerical values are quite different. Therefore, we use the sine and cosine value to represent angles. Finally, a hand joint is represented by 4 elements $(\rho, \cos (\phi), \sin (\phi), z')$. For a 3D pose $\mathbf{Y}$ in the camera coordinates, the whole transformation process is denoted as $\mathcal{T}(\cdot)$: $  \mathbf{h} = \mathcal{T}(\mathbf{Y}) =vec(\mathcal{F} (\mathbf{Y}))  \in \mathbb{R}^{4m},$
where $\mathcal{F}(\cdot)$ denotes the transformation from camera coordinate to the object-oriented cylindrical coordinate, and $vec(\cdot)$ means the flatten operation to form a column vector. This proposed object-oriented cylindrical coordinate is used for the input and output of pose dictionary learning module in all two training phases.

\vspace{-7pt}
\subsection{Phase I: Train Pose Dictionary Learning Module}
We now present a pose dictionary learning module, which will be used to enable unlabeled data to provide supervision signals. The functionality of a pose dictionary learning module is to reconstruct a 3D grasping pose through a trainable pose dictionary. When this pose dictionary is well trained, the reconstruction error of an input pose can reflect the realistic level of the 3D grasping pose. Therefore, given unlabeled data, this pose dictionary can find unrealistic 3D estimation and provide supervision signals for the pose estimation module.

\begin{figure}[t]
    \centering
    \includegraphics[width=6.5cm]{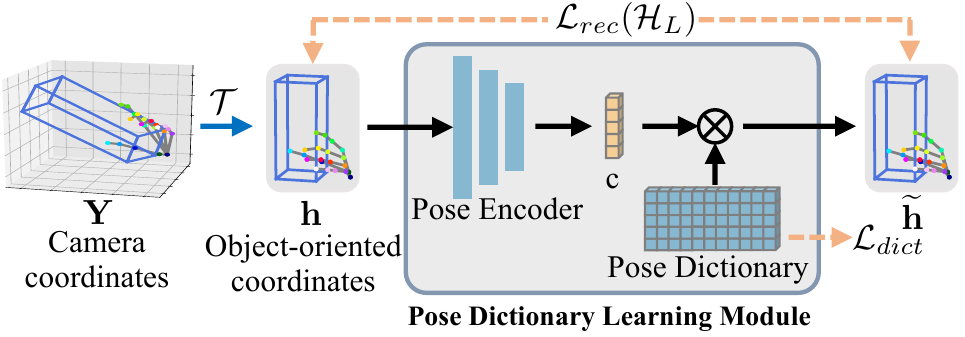}
    \vspace{-15pt}
    \caption{Train the pose dictionary learning module based on labeled data through an auxiliary self-reconstruction task.
    \vspace{-7pt}
    }
    \label{fig:mainimg1}
\end{figure}

We train the pose dictionary learning module through self-reconstruction based on labeled data. As shown in Fig.~\ref{fig:mainimg1}, the pose dictionary learning module is comprised of a trainable pose dictionary $\mathbf{D} \in \mathbb{R}^{4m \times k}$ and a pose encoder $Enc(\cdot, \theta_{e})$, where $k$ is the number of pose atoms and $\theta_{e}$ is the parameters of pose encoder. Each column vector in the pose dictionary is a trainable atom that represents an elementary grasping pose. The pose encoder is based on a multilayer perceptron model with residual connections. In the labeled data, for a 3D pose $\mathbf{Y}$ , we firstly transform it to the proposed object-oriented cylindrical coordinate: $\mathbf{h}=\mathcal{T}(\mathbf{Y})$. Then we input $\mathbf{h}$ to the pose encoder and get the coefficients $\mathbf{c}$: $\ \mathbf{c} \ = \  Enc(\mathbf{h};\theta_{e})  \in \mathbb{R}^{k}$.

Note that the pose encoder $Enc(\cdot,\theta_e)$ includes a softmax operation so that the sum of all elements of $\mathbf{c}$ equals $1$.  To reconstruct a 3D pose, we can use a linear combination of pose atoms to approximate. Here $\mathbf{c}$ performs as the weight coefficients corresponding to the atoms; that is, $\ \mathbf{\widetilde{h}} \ = \ \mathbf{D} \mathbf{c} \in \mathbb{R}^{4m}$. The reconstruction loss is defined by the mean square error:
\begin{equation}
\vspace{-5pt}
\begin{aligned}
    \mathcal{L}_{rec}(\mathcal{H}_{L}) &= \frac{1}{4m\cdot |\mathcal{H}_{L}|} \sum_{\mathbf{h} \in \mathcal{H}_{L} } \left \| \mathbf{D} \cdot Enc(\mathbf{h};\theta_{e}) - \mathbf{h}\right \|^2 \\
    &= \frac{1}{4m\cdot |\mathcal{H}_{L}|} \sum_{\mathbf{h} \in \mathcal{H}_{L} } \left \| \mathbf{\widetilde{h}} - \mathbf{h} \right \|^2,
\end{aligned}
\label{eq:lossrecons1}
\end{equation}
where $\mathcal{H}_L = \left\{ \mathcal{T}(\mathbf{Y}) | \mathbf{Y} \in \mathcal{Y}_L \right\}$.
    
    
    
\begin{figure}[t]
    \centering
    \includegraphics[width=8.0cm]{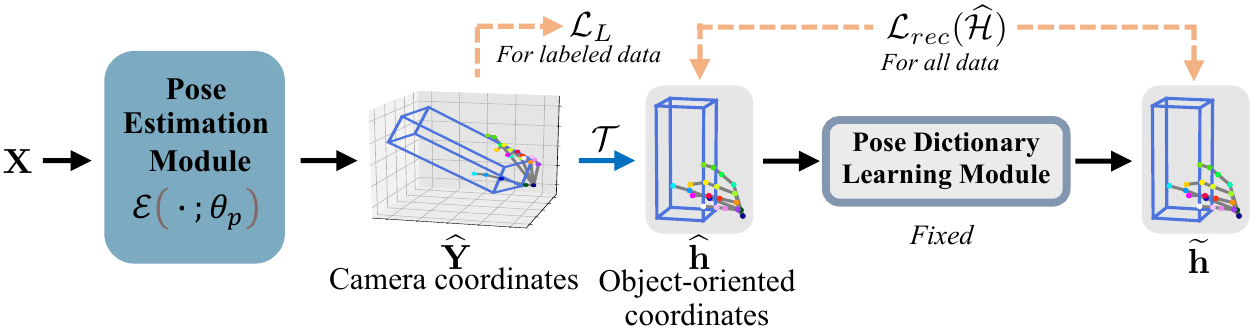}
    \vspace{-15pt}
    \caption{Train the pose estimation module with pose dictionary learning module fixed based on  all data. 
    \vspace{-12pt}
    }
    \label{fig:mainimg2}
\end{figure}

To ease the training of the pose dictionary $\mathbf{D}$, we perform $k$-means clustering on $\mathcal{H}_L$ and initialize $\mathbf{D}$ with those cluster centers. To ensure each pose atom is feasible, we put an additional loss on $\mathbf{D}$ to constrain its elements in valid intervals. Specifically, elements for sine and cosine values should lie in $[-1,1]$ and $\rho$ values should be no less than $0$. Finally we define the valid dictionary loss $\mathcal{L}_{dict}$ by: 
\begin{equation}
\begin{aligned}
    \mathcal{L}_{dict} = \frac{2}{3|\mathcal{D}_{sc}|} \sum_{d \in \mathcal{D}_{sc}} \mathcal{I}(d ; -1, 1)
    + \frac{1}{3|\mathcal{D}_{\rho}|} \sum_{d \in \mathcal{D}_{\rho}} \mathcal{I}(d ; 0, +\infty),
\end{aligned}
\end{equation}
where $\mathcal{D}_{sc}$ is the set of sine and cosine elements in $\mathbf{D}$, $\mathcal{D}_{\rho}$ is the set of $\rho$ elements, and the interval loss function $\mathcal{I}(\cdot)$ is defined as $\mathcal{I}(d ; d_{min}, d_{max})=\max (d_{min}-d, 0)+\max (d-d_{max}, 0)$.

The final loss for training the pose dictionary learning module is: $\ \mathcal{L}_{pdl} = \mathcal{L}_{rec}(\mathcal{H}_{L}) + \lambda_{dict} \mathcal{L}_{dict}$,
where $\lambda_{dict}$ is the weight for valid dictionary loss and we empirically set it to a large value to make sure pose atoms are valid.

To sum up, the pose dictionary learning module is comprised of a trainable pose dictionary and a pose encoder. It performs the reconstruction of 3D poses while the reconstruction process is regularized by the pose dictionary. By training the pose dictionary learning module on the limited annotated data, we squeeze information about feasible grasping poses into the pose dictionary. After the training is finished, the reconstruction error of a given grasping pose can reflect whether the input is realistic.

\subsection{Phase II: Train Pose Estimation Module}
The pose estimation module $\mathcal{E}(\cdot;\theta_{p})$ is implemented based on an adaptive graph U-net proposed by \cite{HOPE} with parameters $\theta_{p}$. It takes a 2D pose $\mathbf{X}$ as input and produces an estimated 3D pose, $\widehat{\mathbf{Y}} = \mathcal{E}(\mathbf{X};\theta_{p})$.

To train this network, we consider supervisions from two aspects; see an overall illustration in Fig.~\ref{fig:mainimg2}.
First, based on labeled data, we consider a direct fully-supervised estimation loss (in the camera coordinate $xyz$):
\begin{equation}
\begin{aligned}
    \mathcal{L}_{L} = \frac{1}{3(m+8)\cdot N_L} \sum_{(\mathbf{X}, \mathbf{Y}) \in \mathcal{P}_L} \left \| \mathcal{E}(\mathbf{X};\theta_{p}) - \mathbf{Y} \right \|_{\rm F}^2,
\end{aligned}
\label{eq:losssfullsuper}
\end{equation}
where $|\mathcal{P}_L| = N_L$ is the number of labeled data and $\left \|\cdot \right \|_{\rm F}$ means Frobenius norm.

Second, based on both labeled and unlabeled data, we consider a self-supervision loss. We input all data to the pose estimation module and obtain 3D estimations; we next represent those estimations in the proposed object-oriented cylindrical coordinate; finally, we input the transformed estimations to the fixed pose dictionary learning module. Recall the definition of reconstruction loss in the first training phase in Eq.~\eqref{eq:lossrecons1}. Simply changing $\mathcal{H}_{L}$ to $\widehat{\mathcal{H}}$, the reconstruction loss in the second training phase is:  $\mathcal{L}_{rec}(\widehat{\mathcal{H}})$, where $\widehat{\mathcal{H}} = \left \{ \mathcal{T} (\mathcal{E}(\mathbf{X};\theta_{p})) | \mathbf{X} \in \mathcal{X} \right \}$.

Here we freeze the parameters of the pose dictionary learning module and only update $\theta_{p}$. When the reconstruction loss is large, it reflects that the estimation result from the pose estimation module is unrealistic and the well-trained pose dictionary learning module cannot represent such 3D poses. Therefore, minimizing the reconstruction loss pushes the the pose estimation network to produce realistic estimations. The reconstruction loss enables non-annotated data $\mathcal{X}_U$ to provide supervision signals.

The overall loss to train the pose estimation module is:
\begin{equation}
\label{eq:lossstage2}
    \mathcal{L} = \mathcal{L}_L + \lambda_r \mathcal{L}_{rec}(\widehat{\mathcal{H}}),
\end{equation}
where $\lambda_r$ balances the two terms.

\section{EXPERIMENTS}
\label{sec:pagestyle}
\subsection{Datasets and Implementation Details}
\textbf{First-Person Hand Action Dataset(FPHA)}~\cite{FPHA} contains first-person videos of hands manipulating various objects. A subset of it (21501 frames) include 3D annotations of 21 hand joints and 8 object bounding box corners. We remove the frames where the hand is not in contact with the object. Finally we get 10388 frames for training and 9761 for evaluation. \textbf{HO-3D}~\cite{HO3D} dataset contains videos of hand-object interaction from third-person view. Here we use the subset for seen-object setting and remove frames without contact. As the frame rate is high, we take images every other frame in the training set. We finally use 9467 frames for training and 6120 for evaluation. For hand annotations, the evaluation set only provides the wrist joints so we only evaluate on wrist joints. HO-3D dataset is of low diversity, and some \emph{different} video sequences are actually the \emph{same} period of action captured by multiple cameras from different perspectives. Thus, the results of HO-3D is less convincing so we mainly demonstrate the results of FPHA. 

For both datasets, we sample 5\% of the frames as the annotated samples $\mathcal{X}_L $. As our method is based on single image, temporal information of video data can affect our evaluation of the model's performance. To reduce the affection, we divide the videos into subsequences of 5 frames each and randomly sample 5\% of the subsequences as $\mathcal{X}_L$. The pose encoder is an 8-layer MLP model with (1024,256,256,1024,256,256,1024,$k$) hidden units in each layer. Two shortcut connections are added between output of (1st,4th) and (4th,7th) layers, similar to  \cite{c3dpo}. ReLU activation and batch normalization are used after every layer except the last. We set $\lambda_{dict}$ and $\lambda_r$ to $100$ and the size $k$ of pose dictionary is 30 by default. As semi-supervised task is sensitive to the amount of data, we do not perform pretraining on large scale synthetic dataset~\cite{Obman}, which is different from \cite{HOPE,HOT}.

\subsection{Results}
\textbf{Primary results}: Table \ref{tab:main} reports the results on FPHA and HO-3D where Procrustes Aligned Mean Per Joint Position Error (MPJPE) is used as the metric. Fig.~5 reports the Percentage of Correct Keypoints (PCK) on FPHA. 
We compare with following methods: (1) training the pose estimation module with all samples labeled (fully supervision), which is the error lower bound; (2) training the pose estimation module only by the 5\% labeled data straightforwardly (5\% supervision), which is the error upper bound; (3) Biomechanical Constraints (BMC)~\cite{BMC} method transplanted to our task; (4) replacing the pose dictionary learning module with a vanilla autoencoder (AE reconstructor), which also performs the reconstruction task. Compared to straightforward use of the limited labeled data, our method significantly reduces the error. The proposed method outperforms BMC on FPHA and acquires comparative results on HO-3D. (HO-3D is less suitable for this task and its results is less convincing.) Our method surpasses AE reconstructor, validating the effectiveness of pose dictionary learning.
As the annotated data $\mathcal{X}_L$ is selected from video sequences, we study the affection of temporal information. We create pseudo-labels for $\mathcal{X}_U$ by interpolation along time and train the pose estimation module by both original and pseudo-labels. Finally we get dramatically large error ($>$25mm/50mm) on hands/objects. It suggests that the affection of temporal information is eliminated.

\begin{table}[t]
\centering
\caption{MPJPE(mm) on FPHA and HO-3D. Best results in bold.}
\label{tab:main}
    \begin{tabular}{crrrr}
    \toprule
    \multirow{2}*{Method}            & \multicolumn{2}{c}{FPHA}        & \multicolumn{2}{c}{HO-3D}       \\ \cline{2-5}
     ~ & Hand & Object & Hand & Object \\ \hline
    Fully supervision & 8.77  & 14.31          & 54.94          & 32.17          \\
    5\% supervision   & 14.00          & 25.19          & 59.09          & 36.12          \\
    BMC~\cite{BMC}               & 11.67          & 21.99          & 56.24          & \textbf{33.24} \\
    AE reconstructor        & 11.92          & 21.53          & 58.36          & 34.80 \\
    Ours              & \textbf{11.27} & \textbf{18.91} & \textbf{56.09} & 33.58          \\
    \bottomrule
    \end{tabular}
    \vspace{-13pt}
\end{table}

\begin{figure}[t]
\begin{floatrow}[2]
\figurebox{\centerline{\caption{PCK on FPHA }}
  \label{fig:PCK}}{%
  \includegraphics[width=3.5cm]{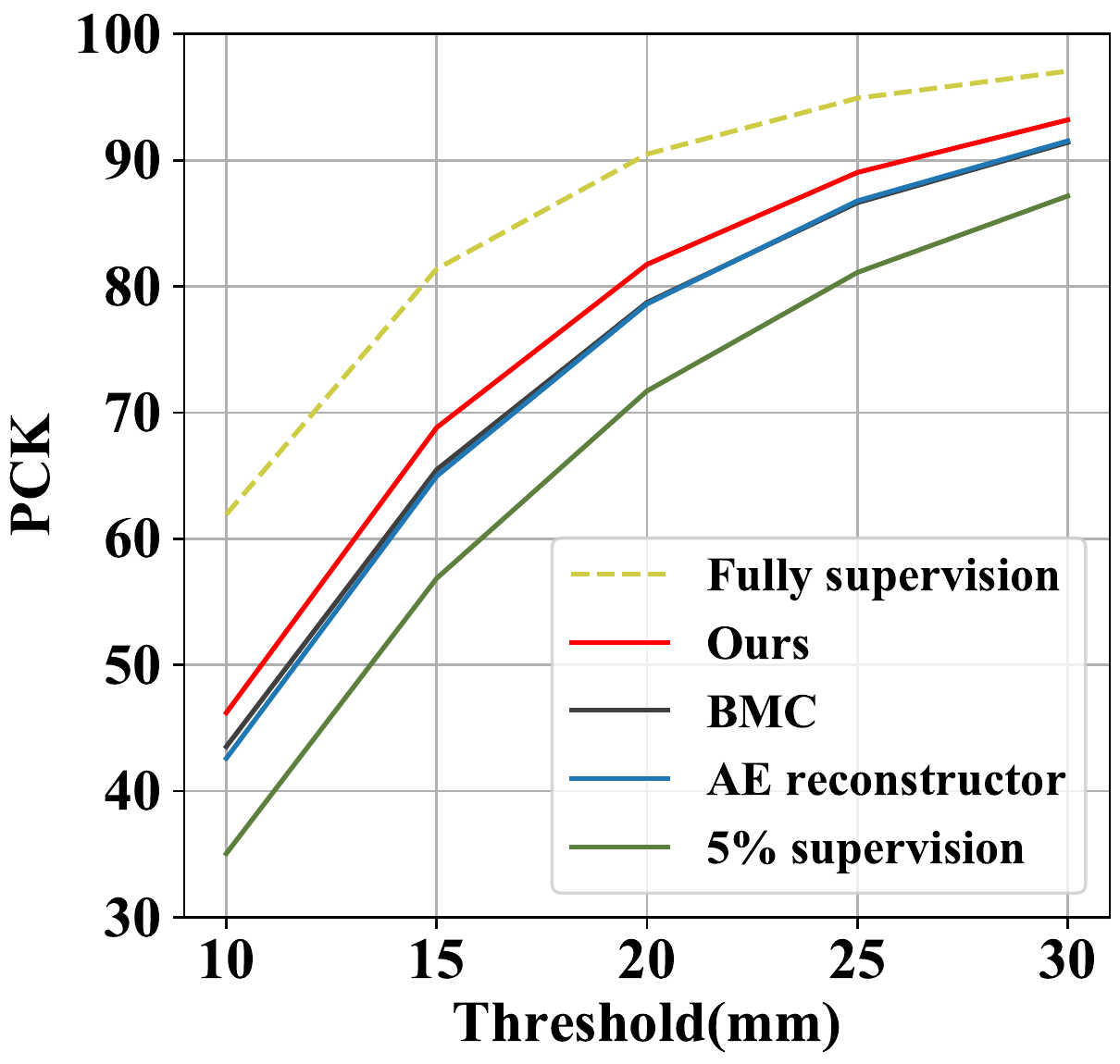}
  \vspace{-15pt}
  }%
  
\hfill
\tablebox{\caption{Impact of $\mathcal{L}_{dict}$}}{%
    \begin{tabular}{ccc}
    \toprule
    Method         & Hand           & Object         \\ \hline
    Ours           & \textbf{11.27} & \textbf{18.91} \\
    L2            & 11.35          & 20.33    \\
    \bottomrule
    \end{tabular}
    \label{tab:lossbasis}
  }
\end{floatrow}
\vspace{-8pt}
\end{figure}

\begin{figure}[t]
    \centering
    \includegraphics[width=6.0cm]{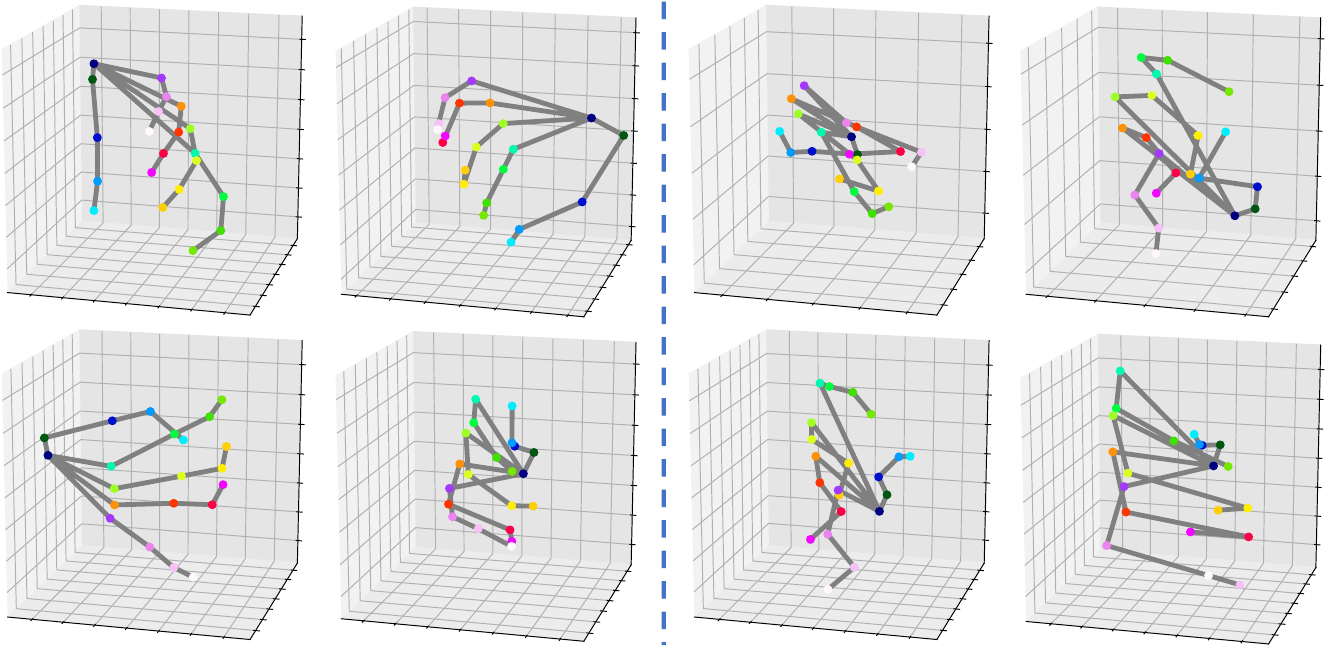}
    \vspace{-19pt}
    \caption{Visualized examples of the learned pose dictionary. Left: our method; right: replacing $\mathcal{L}_{dict}$ with $L2$ regularization. Our method learns a more feasible pose dictionary.}
    \label{fig:basisimg}
\end{figure}

\textbf{Constraints on Pose Dictionary}: Table~2 studies the impact of the constraints on pose dictionary $\mathcal{L}_{dict}$ on FPHA dataset. We replace $\mathcal{L}_{dict}$ with simple $L2$ regularization, which slightly increases the errors. Therefore, we believe that constraining elements of $\mathbf{D}$ in valid intervals can give feasible pose atoms and model the grasping poses better. In Fig.~\ref{fig:basisimg}, we visualize several atom vectors from learned pose dictionary $\mathbf{D}$ with $\mathcal{L}_{dict}$ or $L2$ regularization. The results show that our method learns a more feasible pose dictionary.

\begin{figure}[t]
\begin{minipage}[t]{0.48\linewidth}
  \centering
  \centerline{\includegraphics[width=4.0cm]{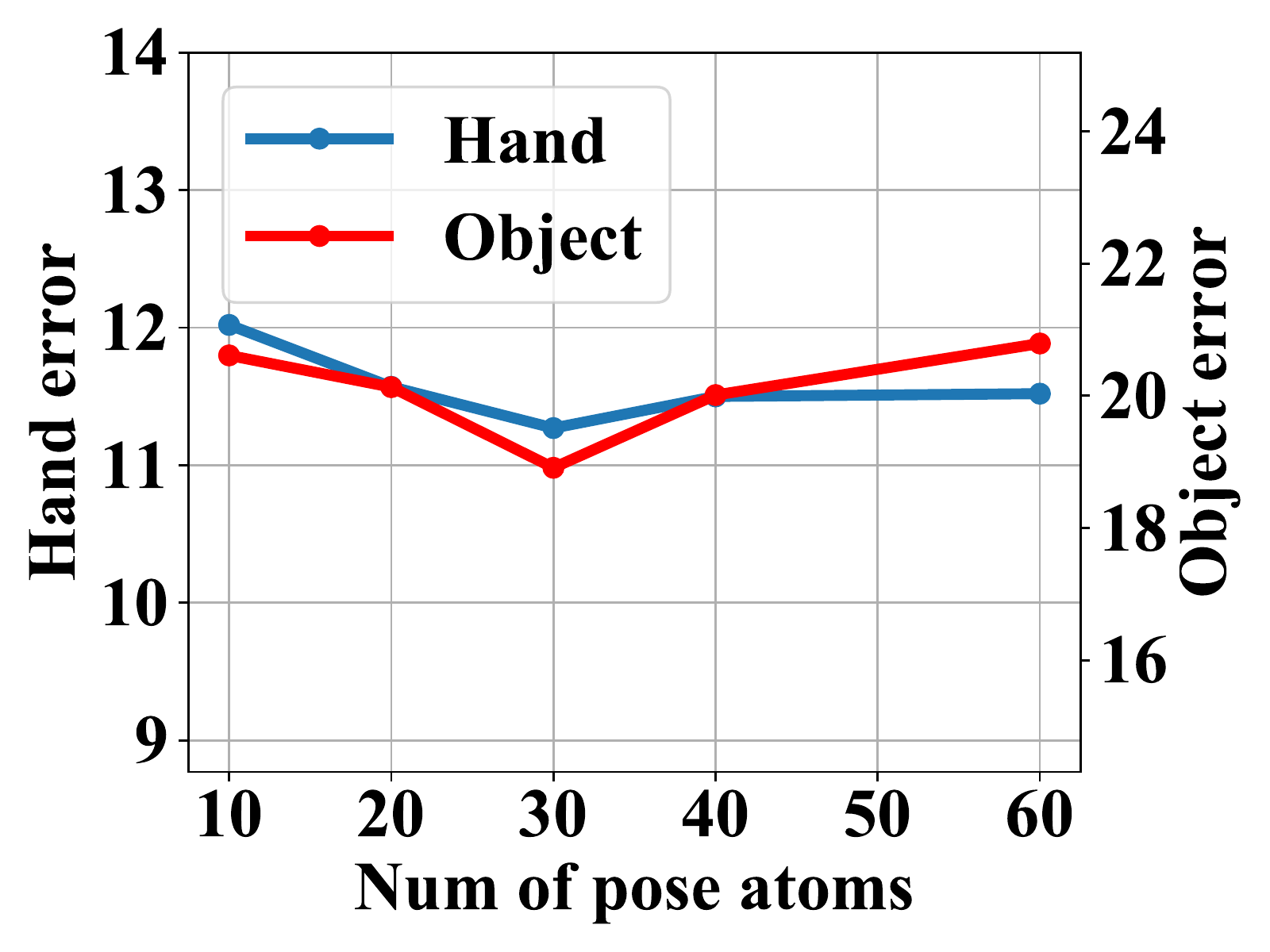}
  \vspace{-4pt}
  }
  \centerline{(a) Impact of $k$}\medskip
\end{minipage}
\vspace{-10pt}
\hfill
\begin{minipage}[t]{0.48\linewidth}
  \centering
  \centerline{\includegraphics[width=4.0cm]{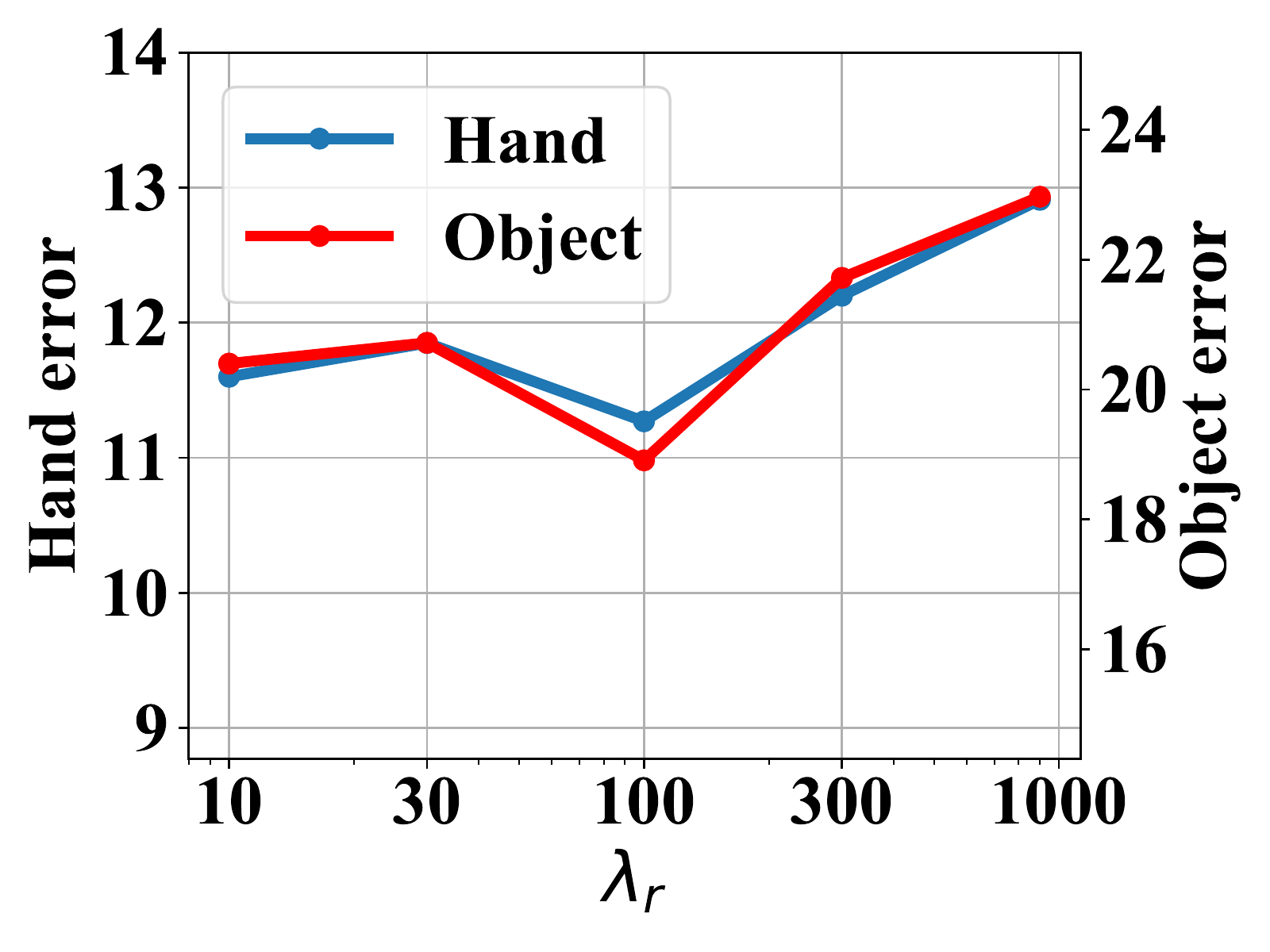}
  \vspace{-4pt}
  }
  \centerline{(b) Impact of $\lambda_r$}\medskip
\end{minipage}
\vspace{-10pt}
\caption{Impact of $k$ and $\lambda_r$. Our method shows robustness to choice of the two hyperparameters.
\vspace{-10pt}
}
\label{fig:robustness}
\end{figure}

\textbf{Robustness to hyperparameters}: We study the robustness of the proposed method with respect to important hyperparameters on FPHA dataset. $k$, the number of pose atoms in $\mathbf{D}$, makes direct and significant impact on the reconstruction process in the pose dictionary learning module. The pose dictionary helps model the grasping poses more explicitly, which can be seen as regularization effect. The regularization is strong when $k$ is small. If $k$ is too large, the pose dictionary learning module gains high variance and may not be able to constrain the estimation result effectively. Fig. \ref{fig:robustness}~(a) reports the impact of $k$. We set $k$ in the moderate interval [10,60]. Our method has achieved low estimation errors in this interval, showing robustness to the choice of $k$. Another important hyperparameter is $\lambda_r$, which controls the weight of reconstruction loss in the second training phase. As shown in Fig.~\ref{fig:robustness} (b), our method performs well when $\lambda_r \in [10,100]$. The errors increase when the weight is too large. We believe this is because the pose dictionary learning module is not a perfect reconstructor. An overlarge weight for reconstruction loss forces the pose estimation module to approach an imperfect target too much.

\begin{figure}[h]
    \vspace{-15pt}
    \centering
    \includegraphics[width=4.5cm]{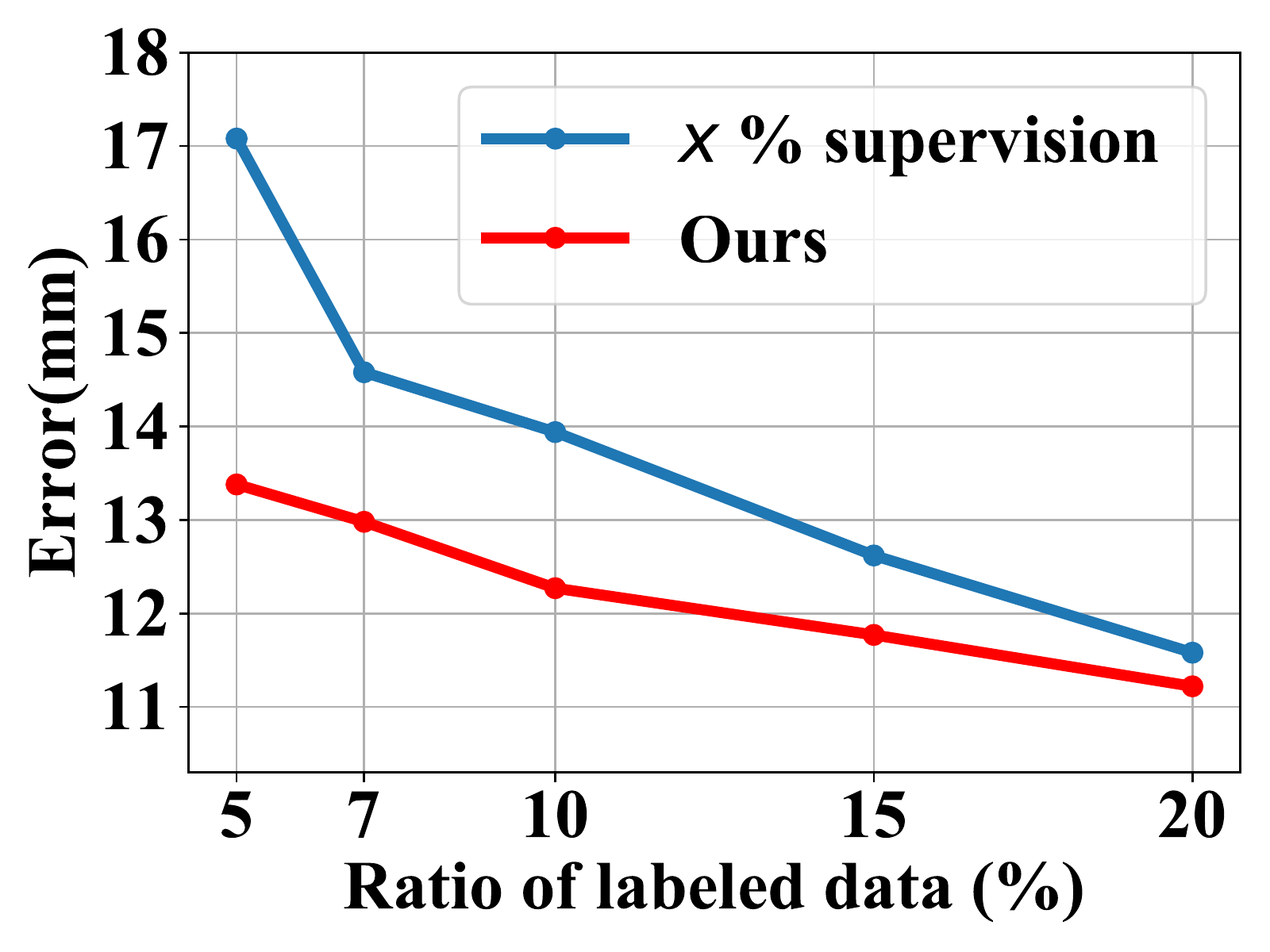}
    \vspace{-15pt}
    \caption{MPJPE of all points (hands and objects) as a function of the labeling ratio on FPHA. Lower bound of y axis is the error of fully supervision. Our method (in red) acquires stable accuracy gain over the straightforward use of labeled data (in blue).\vspace{-20pt}}
    \label{fig:dataratio}
\end{figure}

\textbf{Ratio of labeled data}: The ratio of labeled data is an important factor and we study its impact on estimation accuracy. Fig.~\ref{fig:dataratio} reports the MPJPE of all points. The proposed method (in red) is compared with the straightforward use of labeled data (in blue) on various ratios. The result shows our method can acquire stable accuracy gain.

\section{CONCLUSION}
For hand-object pose estimation task, detailed 3D labels are expensive and labor-intensive. To tackle the data collection problem, we propose a semi-supervised method based on pose dictionary learning. The proposed pose dictionary learning module performs an auxiliary reconstruction task. It enables unlabeled data to provide supervision signals for the pose estimation module by the reconstruction error. We propose to use an object-oriented coordinate system to make the estimation equivariant to the camera perspective. We show that the proposed method improves estimation accuracy significantly and is of high robustness.


\bibliographystyle{IEEEbib}
\bibliography{strings,refs}

\begin{thebibliography}{10}

\bibitem{FPHAafford}
S.~M. {Hussain.S}, L.~{Liu}, W.~{Xu}, and C.~{Lu},
\newblock ``Fpha-afford: A domain-specific benchmark dataset for occluded
  object affordance estimation in human-object-robot interaction,''
\newblock in {\em 2020 IEEE International Conference on Image Processing
  (ICIP)}, 2020, pp. 1416--1420.

\bibitem{handpose_review}
Ammar Ahmad, Cyrille Migniot, and Albert Dipanda,
\newblock ``Hand pose estimation and tracking in real and virtual interaction:a
  review,''
\newblock {\em Image and Vision Computing}, vol. 89, pp. 35 -- 49, 2019.

\bibitem{handpose_app}
Min-Yu Wu, Pai-Wen Ting, Ya-Hui Tang, En-Te Chou, and Li-Chen Fu,
\newblock ``Hand pose estimation in object-interaction based on deep learning
  for virtual reality applications,''
\newblock {\em Journal of Visual Communication and Image Representation}, vol.
  70, pp. 102802, 2020.

\bibitem{handseg_vr}
B.~{Kang}, K.~{Tan}, N.~{Jiang}, H.~{Tai}, D.~{Tretter}, and T.~{Nguyen},
\newblock ``Hand segmentation for hand-object interaction from depth map,''
\newblock in {\em 2017 IEEE Global Conference on Signal and Information
  Processing (GlobalSIP)}, 2017, pp. 259--263.

\bibitem{Tekin_2019_CVPR}
Bugra Tekin, Federica Bogo, and Marc Pollefeys,
\newblock ``H+o: Unified egocentric recognition of 3d hand-object poses and
  interactions,''
\newblock in {\em Proceedings of the IEEE/CVF Conference on Computer Vision and
  Pattern Recognition (CVPR)}, June 2019.

\bibitem{Obman}
Yana Hasson, Gul Varol, Dimitrios Tzionas, Igor Kalevatykh, Michael~J. Black,
  Ivan Laptev, and Cordelia Schmid,
\newblock ``Learning joint reconstruction of hands and manipulated objects,''
\newblock in {\em Proceedings of the IEEE/CVF Conference on Computer Vision and
  Pattern Recognition (CVPR)}, June 2019.

\bibitem{HO_seperate}
D.~{Goudie} and A.~{Galata},
\newblock ``3d hand-object pose estimation from depth with convolutional neural
  networks,''
\newblock in {\em 2017 12th IEEE International Conference on Automatic Face
  Gesture Recognition (FG 2017)}, 2017, pp. 406--413.

\bibitem{HOPE}
Bardia Doosti, Shujon Naha, Majid Mirbagheri, and David~J. Crandall,
\newblock ``Hope-net: A graph-based model for hand-object pose estimation,''
\newblock in {\em Proceedings of the IEEE/CVF Conference on Computer Vision and
  Pattern Recognition (CVPR)}, June 2020.

\bibitem{HOT}
Lin Huang, Jianchao Tan, Jingjing Meng, Ji~Liu, and Junsong Yuan,
\newblock ``Hot-net: Non-autoregressive transformer for 3d hand-object pose
  estimation,''
\newblock in {\em Proceedings of the 28th ACM International Conference on
  Multimedia}, New York, NY, USA, 2020, MM '20, p. 3136–3145, Association for
  Computing Machinery.

\bibitem{transformer}
Ashish Vaswani, Noam Shazeer, Niki Parmar, Jakob Uszkoreit, Llion Jones,
  Aidan~N Gomez, {\L}ukasz Kaiser, and Illia Polosukhin,
\newblock ``Attention is all you need,''
\newblock in {\em Advances in Neural Information Processing Systems}, 2017, pp.
  5998--6008.

\bibitem{FPHA}
Guillermo Garcia-Hernando, Shanxin Yuan, Seungryul Baek, and Tae-Kyun Kim,
\newblock ``First-person hand action benchmark with rgb-d videos and 3d hand
  pose annotations,''
\newblock in {\em Proceedings of the IEEE/CVF Conference on Computer Vision and
  Pattern Recognition (CVPR)}, June 2018.

\bibitem{HO3D}
Shreyas Hampali, Mahdi Rad, Markus Oberweger, and Vincent Lepetit,
\newblock ``Honnotate: A method for 3d annotation of hand and object poses,''
\newblock in {\em Proceedings of the IEEE/CVF Conference on Computer Vision and
  Pattern Recognition (CVPR)}, June 2020.

\bibitem{contactpose}
Brahmbhatt Samarth, Tang Chengcheng, D.~Twigg Christopher, C.~Kemp Charles, and
  Hays James,
\newblock ``Contactpose: A dataset of grasps with object contact and hand
  pose,''
\newblock in {\em European Conference on Computer Vision (ECCV)}, August 2020.

\bibitem{AE_anomaly_detection}
Manass{\'e}s Ribeiro, Andr{\'e}~Eug{\^e}nio Lazzaretti, and Heitor~Silv{\'e}rio
  Lopes,
\newblock ``A study of deep convolutional auto-encoders for anomaly detection
  in videos,''
\newblock {\em Pattern Recognition Letters}, vol. 105, pp. 13--22, 2018.

\bibitem{AE_image_compression}
Lucas Theis, Wenzhe Shi, Andrew Cunningham, and Ferenc Husz{\'{a}}r,
\newblock ``Lossy image compression with compressive autoencoders,''
\newblock in {\em 5th International Conference on Learning Representations,
  {ICLR} 2017, Toulon, France, April 24-26, 2017, Conference Track
  Proceedings}. 2017, OpenReview.net.

\bibitem{AE_feature}
Jonathan Masci, Ueli Meier, Dan Cire{\c{s}}an, and J{\"u}rgen Schmidhuber,
\newblock ``Stacked convolutional auto-encoders for hierarchical feature
  extraction,''
\newblock in {\em Artificial Neural Networks and Machine Learning -- ICANN
  2011}, Timo Honkela, W{\l}odzis{\l}aw Duch, Mark Girolami, and Samuel Kaski,
  Eds., Berlin, Heidelberg, 2011, pp. 52--59, Springer Berlin Heidelberg.

\bibitem{AE_denoising}
Pascal Vincent, Hugo Larochelle, Yoshua Bengio, and Pierre-Antoine Manzagol,
\newblock ``Extracting and composing robust features with denoising
  autoencoders,''
\newblock in {\em Proceedings of the 25th international conference on Machine
  learning}, 2008, pp. 1096--1103.

\bibitem{HO_unknownobj}
Chiho Choi, Sang Ho~Yoon, Chin-Ning Chen, and Karthik Ramani,
\newblock ``Robust hand pose estimation during the interaction with an unknown
  object,''
\newblock in {\em Proceedings of the IEEE/CVF International Conference on
  Computer Vision (ICCV)}, Oct 2017.

\bibitem{HO_feedback}
M.~{Oberweger}, P.~{Wohlhart}, and V.~{Lepetit},
\newblock ``Generalized feedback loop for joint hand-object pose estimation,''
\newblock {\em IEEE Transactions on Pattern Analysis and Machine Intelligence},
  vol. 42, no. 8, pp. 1898--1912, 2020.

\bibitem{c3dpo}
David Novotny, Nikhila Ravi, Benjamin Graham, Natalia Neverova, and Andrea
  Vedaldi,
\newblock ``C3dpo: Canonical 3d pose networks for non-rigid structure from
  motion,''
\newblock in {\em Proceedings of the IEEE/CVF International Conference on
  Computer Vision (ICCV)}, October 2019.

\bibitem{BMC}
Spurr Adrian, Iqbal Umar, Molchanov Pavlo, Hilliges Otmar, and Kautertz Jan,
\newblock ``Weakly supervised 3d hand pose estimation via biomechanical
  constraints,''
\newblock in {\em European Conference on Computer Vision (ECCV)}, August 2020.

\end{thebibliography}

\end{document}